\relax
\documentclass[letterpaper]{article} 
\usepackage{aaai20}  
\usepackage{times}  
\usepackage{helvet} 
\usepackage{courier}  
\usepackage[hyphens]{url}  
\usepackage{graphicx} 
\usepackage{graphicx}  
\usepackage{amsmath,mathtools,amssymb,amsthm}
\usepackage{subcaption}
\usepackage{bbm}
\usepackage[dvipsnames]{xcolor}

\title{RL-Duet: Online Music Accompaniment Generation Using Deep Reinforcement Learning}
\author{Nan Jiang\textsuperscript{\rm 1} Sheng Jin\textsuperscript{\rm 1}  Zhiyao Duan\textsuperscript{\rm 2}  Changshui Zhang\textsuperscript{\rm 1} \\
\textsuperscript{\rm 1}
Department of Automation, Tsinghua University \\
State Key Lab of Intelligent Technologies and Systems \\
Institute for Artificial Intelligence, Tsinghua University (THUAI)\\ 
Beijing National Research Center for Information Science and Technology (BNRist)\\
\textsuperscript{\rm 2} Department of Electrical and Computer Engineering, University of Rochester \\
\{jiangn15, js17\}@mails.tsinghua.edu.cn, \\zhiyao.duan@rochester.edu, zcs@mail.tsinghua.edu.cn 
}
\begin{document}

\maketitle

\begin{abstract}
This paper presents a deep reinforcement learning algorithm for online accompaniment generation, with potential for real-time interactive human-machine duet improvisation. Different from offline music generation and harmonization, online music accompaniment requires the algorithm to respond to human input and generate the machine counterpart in a sequential order. We cast this as a reinforcement learning problem, where the generation agent learns a policy to generate a musical note (action) based on previously generated context (state). The key of this algorithm is the well-functioning reward model. Instead of defining it using music composition rules, we learn this model from monophonic and polyphonic training data. This model considers the compatibility of the machine-generated note with both the machine-generated context and the human-generated context. Experiments show that this algorithm is able to respond to the human part and generate a melodic, harmonic and diverse machine part. Subjective evaluations on preferences show that the proposed algorithm generates music pieces of higher quality than the baseline method.

\end{abstract}

\section{Introduction}
\label{sec:introduction}

Learning based automatic music generation has been an active research area~\cite{tatar2019musical,briot2017deep}. Existing learning based music generation methods, however, have paid less attention to real-time interactive music generation (\textit{e.g.}, interactive duet improvisation) between humans and machines: A machine agent and a human player collaboratively create music by listening to each other. In this work, we take one step towards this goal and aim to solve the online music accompaniment generation as the first step. Online music accompaniment generation is an interesting but challenging problem. It is interesting because it requires comprehensive musicianship skills including perception and composition, and the ability of fast decision making. It further requires machines to adapt and respond to human input in a timely fashion, leading to successful human-computer collaboration.

\begin{figure}[t]
	\centering
	\includegraphics[width=0.90\columnwidth]{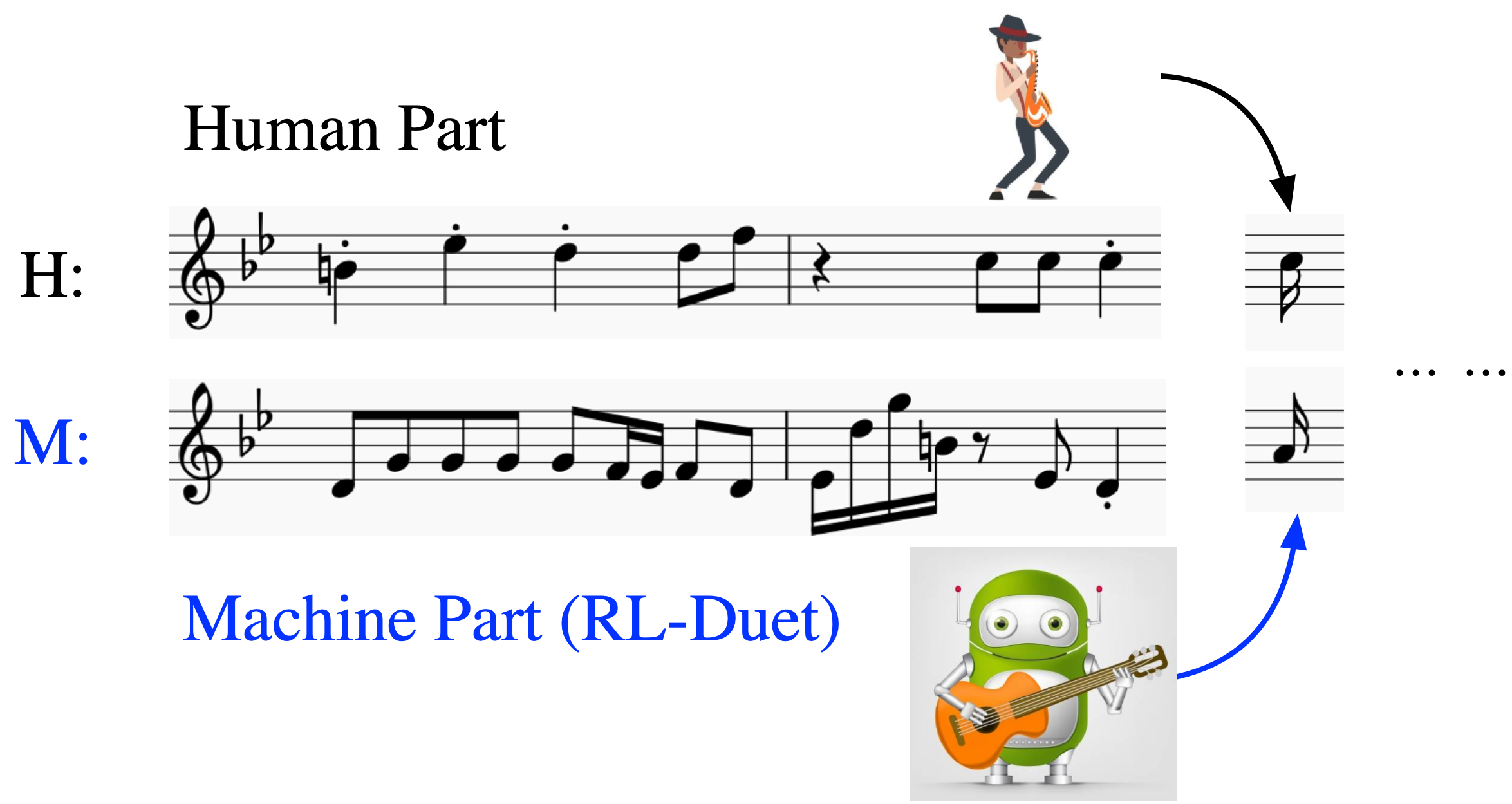}
	\caption{Online music accompaniment generation requires the agent to adapt/respond to human input without delay.}
	\label{fig:intro}
\end{figure}

In this work, we propose a novel algorithm (RL-Duet) for online music accompaniment generation that could be further developed toward a real-time interactive duet improvisation system. In this setting, a human performer would improvise a melody, and the machine would generate a counterpoint accompaniment in real-time. We focus on the design of the algorithm instead of the system; the latter would require significant considerations on the interactivity and usability through systematic user studies, but is beyond the scope of this paper. We cast the problem of online music accompaniment as a reinforcement learning problem. The generation agent takes into consideration both the long-term temporal structure and the inter-part harmonization, and generates musical notes based on the context. 

Reinforcement learning-based music generation has been explored in~\cite{jaques2017sequence}, where hand-crafted composition rules or criteria are used as the reward function for music generation. The main disadvantages of rule-based reward models, we argue, are two-fold: 
1) It is difficult to define a comprehensive list of rules.
There is a long list of composition rules or evaluation criteria in literature~\cite{green1989practical,jaques2017sequence,dong2018musegan,trieu2018jazzgan,yang2018evaluation}. Some of them are hard rules (\textit{e}.\textit{g}. , note duration greater than 32nd note), while others are soft rules (\textit{e.g.}, less empty bars, and more pitch variations). Most of them only focus on one limited aspect, enforcing some constraints on the process of music composition. 
2) It is difficult to balance different rules. 
First, the objective evaluation of music composition is multi-criteria, where good ratings on one criterion does not necessarily imply good ratings on another. In addition, some of the rules can be contradictory to each other. For example, some hand-crafted rules punish flat pitch contours and give high rewards to random note fluctuations, while other rules encourage repetition. 
Finally, the integration and balancing of these rules depend on the music style and vary with music context and progression. The aforementioned problems of rule-based rewards/criteria have also been discussed in other research domains, such as image generation~\cite{theis2016note}.

Instead of laboriously designing rule-based reward functions, we directly learn the reward model from monophonic and polyphonic music. This reward model considers both the inter-part and intra-part compatibility of the generated notes. It avoids unnecessary or incorrect constraints enforced by human specialists. Therefore, it has the potential of generating more melodic and creative melodies. 

To better understand our proposed model, we present qualitative analysis, probing the intriguing nature of our proposed reward model. We also conduct a user study for subjective evaluation, which demonstrates the effectiveness of our proposed method.

Our main contributions are as follows:
\begin{enumerate}
    \item To our best knowledge, RL-Duet is the first reinforcement learning model for online music accompaniment generation using an ensemble of reward models.
    \item Instead of relying on a rule-based objective function, RL-Duet learns a comprehensive reward model from data, considering both the inter-part and intra-part harmonization of human and machine inputs.
    \item Subjective evaluation shows that RL-Duet generates music pieces of higher preferences than the baseline method.
\end{enumerate}

\section{Related Work}

\subsection{Symbolic Music Generation with Deep Learning}
There is a long history of generating music with neural networks \cite{briot2017deep}. while a few works directly generate audio signals \cite{van2016wavenet,engel2018gansynth}, here we focus on symbolic-domain generation. Most works \cite{mozer1994neural,eck2002first,hutchings2017using,zhu2018xiaoice} use RNN (or LSTM, GRU) to generate single-track or multi-track music sequentially. Other works employ the hierarchical structure \cite{chu2016song} or the attention mechanism \cite{waite2016project,huang2019musictransformer} to model long-term temporal dependencies. Generative Adversarial Networks (GAN) have also been used to learn the distribution of music pieces \cite{yang2017midinet,dong2018musegan,mogren2016c,yu2017seqgan}. However, GAN demands a large amount of data and it is hard to stabilize the training process.

Recently, the Gibbs sampling approach \cite{hadjeres2017deepbach,huang2017counterpoint,yan2018part} has shown its effectiveness in generating coherent chorales, since the generation process of Gibbs sampling resembles how humans compose music: Iteratively modifying the music fragments based on the whole music context. However, Gibbs sampling could only be used offline. In this work, we focus on online music accompaniment generation toward real-time human machine interactive improvisation, which requires the generation be online and in a sequential order without delay.

\subsection{Interactive Music Generation Systems}
A recent review~\cite{tatar2019musical} presents a detailed survey of interactive musical agent systems, from rule-based to learning-based systems. According to~\cite{tatar2019musical}, interactive music generation methods can be categorized into \textit{call and response} and \textit{accompaniment}. 

For \textit{call and response}, the agent learns to trade  solos with the user in real-time, i.e., the agent and the user take turns to play. Evolutionary computation-based algorithms~\cite{biles1994genjam,papadopoulos1998genetic} employ a fitness function to perform the genetic selection. Usually, the fitness function relies heavily on the musical knowledge. Rule-based algorithms~\cite{thom2000bob} learn the playing mode of a human performer, and applies percept-to-action functions for solo trading.

This work belongs to online music \textit{accompaniment} generation, which requires the machine agent to adapt and support the human player with accompaniment. In the early Music Information Retrieval (MIR) literature, ``music accompaniment'' often refers to algorithms and systems that are able to render pre-recorded/synthesized accompaniment part following the real-time human solo performance~\cite{dannenberg1984line,raphael2010music}. These systems do not generate or improvise music. In this paper, we focus on accompaniment generation instead of accompaniment matching. Recently, MuseGAN \cite{dong2018musegan} is proposed to use GAN-based models to generate multi-track music. It could be extended to human-AI cooperative music generation. However, it generates additional tracks following a specific track composed by human offline, not designed for interactive human-machine duet improvisation. Google built an interactive Bach Doodle \cite{huang2019bach} to harmonize user created melody. It uses Gibbs sampling to generates the other three parts offline as well.
Our proposed RL-Duet algorithm, on the other hand, responds to the human input and generates accompaniment in an online fashion without any delay. RL-Duet could be integrated to a real-time human-machine interactive improvisation system, such as our other work \cite{benetatosbachduet}. \cite{roberts2016interactive} presents a demo for generating the responses to short melodic calls and bass accompaniment. However, no technical details nor experiments were provided. 

\subsection{Reinforcement Learning for Sequence Generation}

Recently, in the field of natural language processing, many works have employed reinforcement learning (RL) to directly optimize the tasks' evaluation metrics, such as BLEU or ROUGE \cite{ranzato2015sequence,williams1992simple,bahdanau2016actor}. These metrics provide a reasonable way to measure the quality of the generated sequences. For music generation, however, a key difficulty is on the design of rewards that can measure music generation quality. 

For online music generation, maximum likelihood estimation is the most popular training paradigm. Despite the differences in framework, data representation, and model architecture, they all learn to predict the probability of the next token conditioned on the previous tokens and then maximize the likelihood of the whole dataset. 
A few recent works have incorporated reinforcement learning (RL) for music generation \cite{guimaraes2017objective,jaques2017sequence}. RL uses a different training target which considers the long-term discounted reward of the generated music. Among them, SequenceTutor \cite{jaques2017sequence} is mostly related to our RL-Duet. It also uses a trained neural network to give a partial reward, but mostly relies on more than ten hand-crafted musical rules, and therefore suffers from several problems mentioned in the Introduction Section. Moreover, SequenceTutor does not consider the inter-part harmonization between polyphonic parts, which is critical for human-machine interactive improvisation.

\section{RL-Duet}

\subsection{Motivation for Reinforcement Learning}
\label{sec:motivation}
Most existing works on online music generation use Recurrent Neural Networks (RNN) trained by maximum likelihood estimation (MLE)~\cite{bharucha1989modeling,mozer1994neural,eck2002first}. They train RNNs to learn the probability distribution of the next token conditioned on the previous sequence. While such MLE methods have shown decent results, they suffer from two major drawbacks in sequence generation tasks.

\textbf{Mismatch between training and testing conditions (exposure bias):} The model is typically trained to predict the next token conditioned on the past sequence using clean human-composed music. However, during testing, the model has to make predictions based on its previous predictions. Any prior defective prediction will deteriorate the quality of the following generation, and this deterioration may accumulate for long sequences.

\textbf{Mismatch between training criterion and generation objectives:} To maximize the likelihood of the whole dataset, the MLE model minimizes the cross-entropy loss of the token at each time-step (training criterion), whereas the quality of the generated music relates to the global coherence and harmonization of the entire music piece (generation objectives). In other words, when generating a sequence of notes, MLE model chooses a token with the maximum probability at each time-step, and will not take into account its long-term effect on future tokens of the generated sequence. This mismatch is the reason why beam search is often used to improve sequence generation, which is not applicable in the online setting, however.

We propose to cast the music generation problem as a sequential decision-making problem, and use reinforcement learning (RL) to address the above two problems. The agent learns to maximize the long-term discounted reward by trial and error, directly using model predictions at the training time. Second, we learn a comprehensive reward model from data, which captures 1) the global temporal and rhythmic structure and 2) the harmonic inter-dependency between machine and human parts. 

\begin{figure}[tb]
	\centering
	\includegraphics[width=0.90\columnwidth]{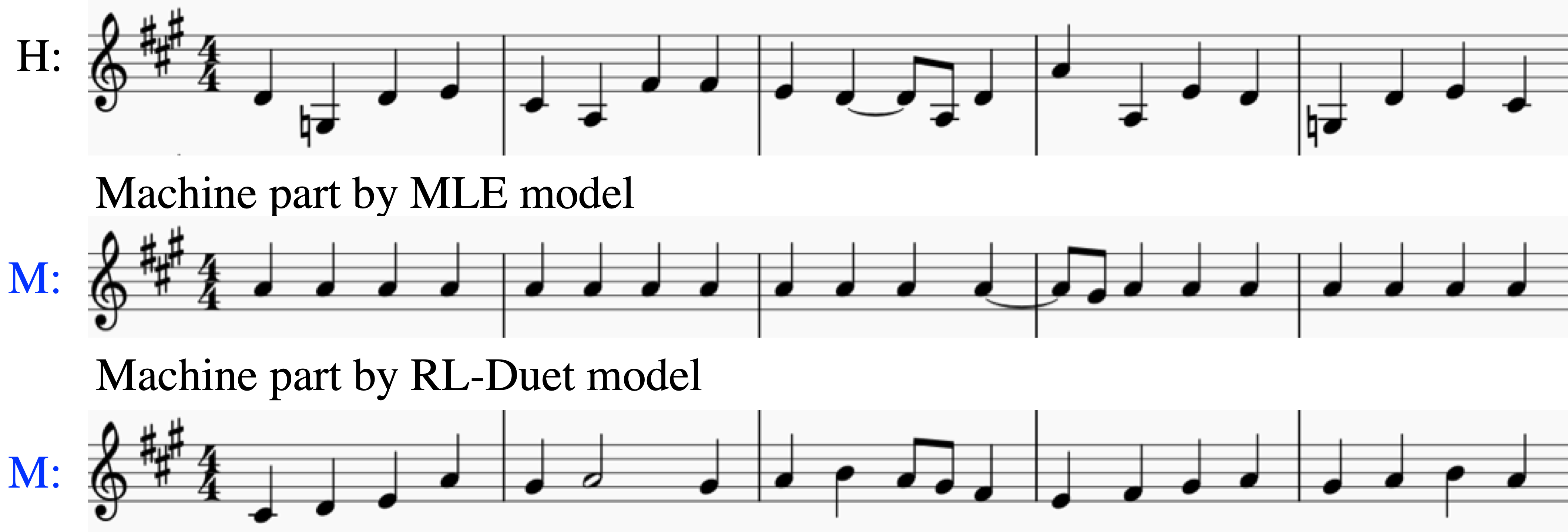}
	\caption{Typical problems of maximum likelihood estimation (MLE) in online accompaniment generation (H: human, M: machine). The MLE model tends to produce many repeated notes, while the proposed RL-Duet model shows better diversity in the melodic contour.
	}
	\label{fig:motivation}
\end{figure}

The MLE model can achieve a high accuracy (83.75\%) of token prediction when it takes a duet composed by Bach as the model input, but when generating on its own, it greedily chooses the note with the highest probability at each time-step, yielding flat and monotonous music. Fig. \ref{fig:motivation} demonstrates this typical failure mode of the MLE model. In comparison, RL-Duet better preserves the temporal and rhythmic structures, diversity and inter-part harmonization.

\begin{figure*}[t]
	\centering
	\includegraphics[width=1.72\columnwidth]{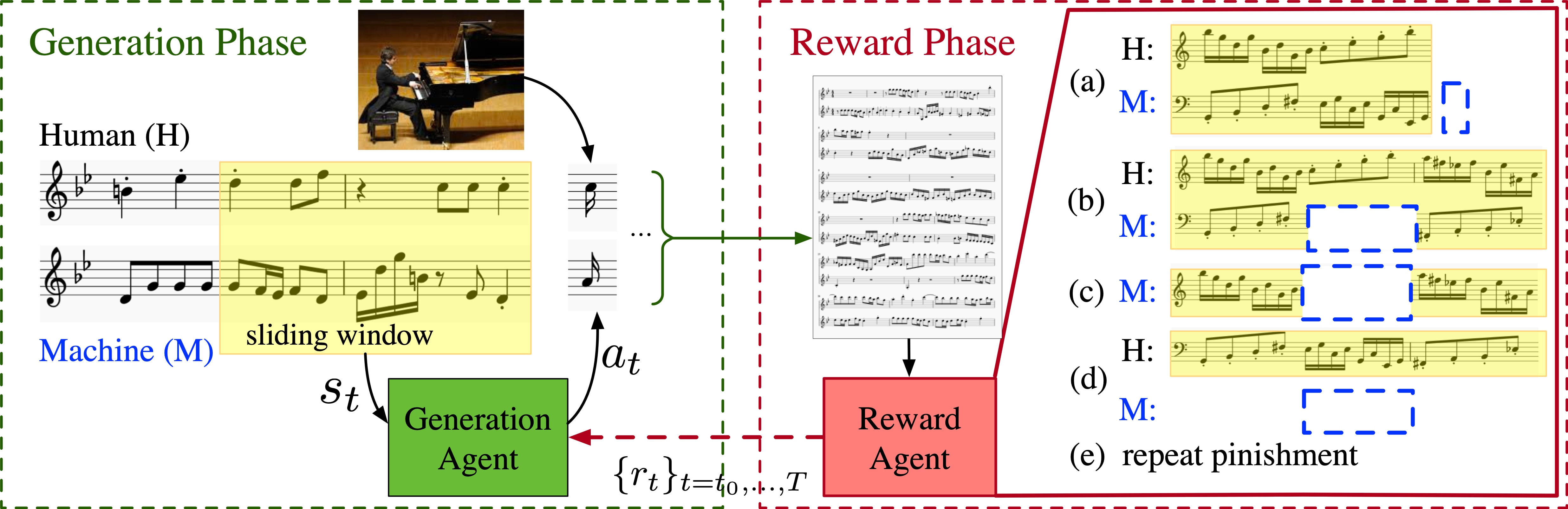}
	\caption{Framework of RL-Duet. \textbf{Generation Phase:} The generation agent sequentially produces notes along with the sequentially received human input. \textbf{Reward Phase:} The reward agent computes a reward for each generated note. It consists of four neural network based submodules and a rule-based reward, which gives a negative punishment if the same pitch is repeated for many times. Each of the submodules is a neural network, which receives the yellow masked area as the input context and outputs the probability (\textit{i}.\textit{e}., reward) of the notes in the dashed blue rectangle region. (a) Joint modeling with pre-context. (b) Joint modeling with both pre- and post- context. (c) Horizontal temporal consistency modeling. (d) Vertical harmonization modeling. 
	}
	\label{fig:framework}
\end{figure*}

\subsection{Representation}
\label{music-rep}
\subsubsection{Pitch, Duration, Articulation}

We use symbolic MIDI pitches to encode notes. We quantize time into sixteenth note, which is the shortest duration of notes in our dataset. For notes with a longer duration, we use ``hold'' symbols to encode their length. Two kinds of pitch representations are used in RL-Duet. One uses a single ``hold'' symbol for all the pitches, the other uses a separate ``hold'' symbol for each pitch. These two representations are referred as ``multi-hold'' and ``single-hold'' respectively. For example, a C4 quarter note will be encoded as four time-steps: [C4, hold, hold, hold] or [C4, C4\_hold, C4\_hold, C4\_hold] in these two representations. The ``single-hold'' representation is common in symbolic music generation literature. However, our experiments show that the ``multi-hold'' representation will generate music of more diversity, as this mitigates the extreme imbalance of the number of the hold symbol and pitch numbers in the single ``hold'' representation. We further encode rests with an additional ``rest'' symbol.

\subsubsection{Beat Information}
Besides the pitch number of each note, the music also contains additional rhythmic information. Following DeepBach \cite{hadjeres2017deepbach}, we make use of a \textit{subdivision} list to encode the beat information. We quantize time with sixteenth notes, and each beat is subdivided into four parts. The subdivision list, $B = \left[b_0, b_1,...,b_T\right]$, containing each time-step's subdivision beat index $b_t$, is a repeated list of $\left[1,2,3,4,1,2,3,4...\right]$. 

\subsubsection{Model Concepts}
An interactive duet improvisation consists of two parts, the human part $H= \left[h_0, h_1,...,h_T\right]$ and the machine part $M= \left[m_0, m_1,...,m_T\right]$, where $h_t$ and $m_t$ are tokens at time-step $t$ in the human part and the machine part, respectively. During improvisation, the state at time-step $t$ is denoted as the joint state of the human part and the machine part, $s_t = (s_{t}^h, s_{t}^m)$. We abbreviate $s_{t}^h = h_{0:t-1}$ and $s_{t}^m = m_{0:t-1}$, representing previous tokens of the human part and the machine part, respectively.

The generation model learns the probability distribution $p(m_t|s_t)$: Given the state at each time-step, it outputs the probability of the next token of the machine part. In our model, we use a sliding window over the previous notes of both parts to represent the state, ignoring the notes far before. Therefore, without ambiguity, the input $s_t$ of our model only contains the clipped pre-tokens, from time-step $t-L$ to $t-1$, if the sliding window size is $L$.

\subsection{RL Framework}
We use the reinforcement learning algorithm, actor-critic with generalized advantage estimator (GAE)~\cite{schulman2015highdimensional}, to train the generation agent. The optimization objective of reinforcement learning is different from that of MLE models. For an MLE model, it trains to learn $p(i|s_t)$ and maximize the likelihood of the whole dataset; During generation, it greedily selects the token with the highest probability as the next token, $m_t = \arg\max_i p(i|s_t)$. With the MLE training criterion and generation policy, their model only focuses on the token-level predictions, ignoring the long-term global consistency and harmonization of the whole music. Reinforcement learning learns a policy that maximizes the expected long-term total reward. 

In contrast, reinforcement learning (RL) optimizes the expected long-term discounted reward $\mathbb{E}_{\pi} R$. The discounted reward at time-step $t$ is $R_t = \sum_{i=0}^{T} \gamma^i r_{i+t}$, where the reward $r_t$ reflects the quality of the generated note $m_t$, and $\gamma$ is the discount factor. We will explain in detail how we obtain a comprehensive reward $r_t$ in the following section. Note that when $\gamma=0, r_t = p(m_t \vert m_{0:t-1}, h_{0:t-1})$, the RL objective only considers the expected reward of a single step, which is the same as that of MLE models. For $\gamma>0$, the model takes the long-term future reward into account. 

In actor-critic with generalized advantage estimator, it learns an action policy $\pi_{\theta_a}(a_t|s_t)$ and a value function $V_{\theta_v}(s_t)$: the agent gives an action $a_t$ (a note token) based on the current state $s_t$ (tokens of the previous time-steps). Here, $a_t$ is equivalent to the $m_t$ in MLE model, and $\pi(\cdot|\cdot)$ to $p(\cdot|\cdot)$ as well.

The discounted policy gradient is 
\begin{align}
g^{\gamma} &\approx \mathbb{E}\left[{\sum_{t=0}^{T} \nabla_{\theta_a} \log \pi_{\theta_a}(a_t \vert s_t) \hat{A}^{\mathrm{GAE(\gamma,\lambda)}}_t} \right] \\ \nonumber
&= \mathbb{E}\left[{\sum_{t=0}^{T} \nabla_{\theta_a} \log \pi_{\theta}(a_t \vert s_t) \sum_{l=0}^{T} (\gamma \lambda)^{l}\delta^V_{t+l}} \right],
\label{eq:pg-gae1}
\end{align}
where $\delta^V_t = r_t + \gamma V(s_{t+1}) - V(s_t)$ is the temporal difference (TD) residual of the value function $V$.

The gradient for the value function is
\begin{align}
 \mathbb{E} \left[\sum_{t=0}^{T} \nabla_{\theta_v} ( \Vert V_{\theta_v} (s_t) - R_t \Vert^2)   \right]. 
\end{align}

The framework of RL-Duet is shown in Fig \ref{fig:framework}. The training process of RL-Duet iterates between two phases: generation phase and reward phase. In the generation phase, the generation agent sequentially produces the counterpart token $a_t$, based on the state $s_t$, generated by both human and machine. After the improvisation of the whole music, the reward agent computes the reward for the action $a_t$ at each time-step. Then these rewards are backtracked to calculate $R_t$ and are used to update the generation agent $\pi_{\theta_a}(a_t|s_t)$ and the reward function $V_{\theta_v}(s_t)$ (not shown in Fig \ref{fig:framework}). 

\subsection{Reward Model}
\label{sec:reward}
Previous RL based models~\cite{jaques2017sequence} rely on music composition rules and heuristics hand-crafted by human specialists. Our algorithm replaces these rules with a set of deep neural networks (reward models) that require no domain knowledge, leading to more creative and diverse melodies.

A well-functioning reward model is key to the success of our algorithm. It is natural to use a single compact reward model, which models $p(m_t \vert s_t)$, to score the music sequences. However, deep neural networks are easily fooled~\cite{nguyen2015deep}. For a single model, there may exist ``undesired regions'' where the reward model assigns unreasonably high scores to unpleasant or boring music excerpts. We call such kind of music sequences the fooling instances. From the perspective of improving the model robustness, \cite{abbasi2017robustness} claim that an ensemble of several specialists will have higher entropy (disagreement) over the scoring of fooling instances, thus is more likely to identify and reject the undesired fooling instances. Motivated by~\cite{abbasi2017robustness}, we design an ensemble of the reward model specialists to avoid these ``undesired regions''. 

Our reward agent considers both the \emph{horizontal} temporal consistency~\cite{briot2017deep} over the machine-generated part, and the \emph{vertical} harmony relations~\cite{briot2017deep} between the human and machine parts. There are four reward models in RL-Duet, which are shown in Fig \ref{fig:framework}. Considering the relative positions of the model's input and output, they could be classified into three types. 

\textbf{Joint Modeling:} 
\label{sec:joint-model}
One way to construct the reward model is to directly learn $p(m_t \vert s_t)$ from the dataset. This reward model contains the same input and output of the generation model, as shown in Fig. \ref{fig:framework}(a). However, this model only tries to capture the relation between the $t$-th token and the pre-context. We improve it by jointly considering both the pre- and post- contexts, $p(m_{t:t+\Delta} \vert \Tilde{s_t})$, as shown in Fig. \ref{fig:framework}(b). We choose to model the probability of a range of notes, from $m_t$ to $m_{t+\Delta}$, for better structured learning. Related to $s_t$, the pre-contextual notes in the human and machine parts, $\Tilde{s_t}$ is the augmented pre- and post- contextual notes. Similarly, we use a sliding window to preserve only the local music excerpts around the tokens $m_{t:t+\Delta}$. 

In order to explicitly model the horizontal and vertical view, we propose to disentangle the joint modeling reward model into two parts.

\textbf{Horizontal View:} 
Another reward model learns to capture the intra-part temporal consistency, incorporating both the pre- and post- contexts of only the machine part. As shown in Fig. \ref{fig:framework}(c), the model is trained with the unsupervised self-prediction Cloze task~\cite{taylor1953cloze,devlin2018bert}, aiming to model the joint probability of the central masked tokens, $p(m_{t:t+\Delta} \vert \Tilde{s}^m_t)$, given its pre- and post- contextual notes. Similar to the $\Tilde{s_t}$ in Joint Modeling reward models, $\Tilde{s}^m_t$ is the augmented pre- and post- contextual notes of the machine part. Moreover, this disentanglement enables the utilization of massive monophonic music sequences to aid the training of this reward model.

\textbf{Vertical View:}
Fig. \ref{fig:framework}(d) shows the last reward model, which is designed to capture the inter-part harmony, $p(m_{t:t+\Delta} \vert \Tilde{s}^h_t)$. Likewise, $\Tilde{s}^h_t$ is the augmented pre- and post- contextual notes of the human part. This reward model ignores the pre- and post context of the machine, only modeling the harmonization of the central excerpt of the machine part and the whole human part.

These four types of reward models serve a comprehensive purpose of achieving inter-part and intra-part coherence. Rewards given by all the reward models are averaged to give a model-based reward. In practice, a user could tune their weights based on his/her preference and experience. Besides, we augment the model-based reward with a simple rule-based reward, a negative punishment, -1, when a note is excessively repeated, adapted from SequenceTutor \cite{jaques2017sequence}. The total reward at each time-step is the sum of the model-based reward and the rule-based reward.

\section{Experiments}

\subsection{Experimental Setup}

\subsubsection{Datasets}

The model is trained on the Bach Chorale dataset in Music21~\cite{music21}. We use chorales with four monophonic parts, in the SATB format (Soprano, Alto, Tenor, and Bass) as the training and validation datasets, with 327 and 37 chorales respectively. When training reward models, the human part and the machine part are randomly chosen from four parts of one chorale to form a duet. We perform data augmentation by transposing the chorales such that the transposed ones do not exceed the highest and the lowest pitches of the original dataset. The MIDI pitches lie in the range from MIDI number 36 to 81. 

When RL-Duet is eventually deployed in a real-time human-machine interactive improvisation system, and the human improvisation would also be influenced by the machine generation. As this paper focuses on the algorithm design and such interactive system is not in place yet, for our subjective evaluation, the human part is drawn from the remaining 37 Bach chorales not used in training and validation. We form 460 duet pairs from these chorales as our test data.
For each duet, the human part is fixed, so is the first two measures of the machine part. This provides an initialization for the generation models to generate the rest of the machine part to accompany the human part. Each generation model greedily generates one accompaniment part for the human part. 
The objective and subjective evaluations of each generation model are based on the 460 generated duets.

\begin{figure}[ht!]
	\centering
	\includegraphics[width=0.95\columnwidth]{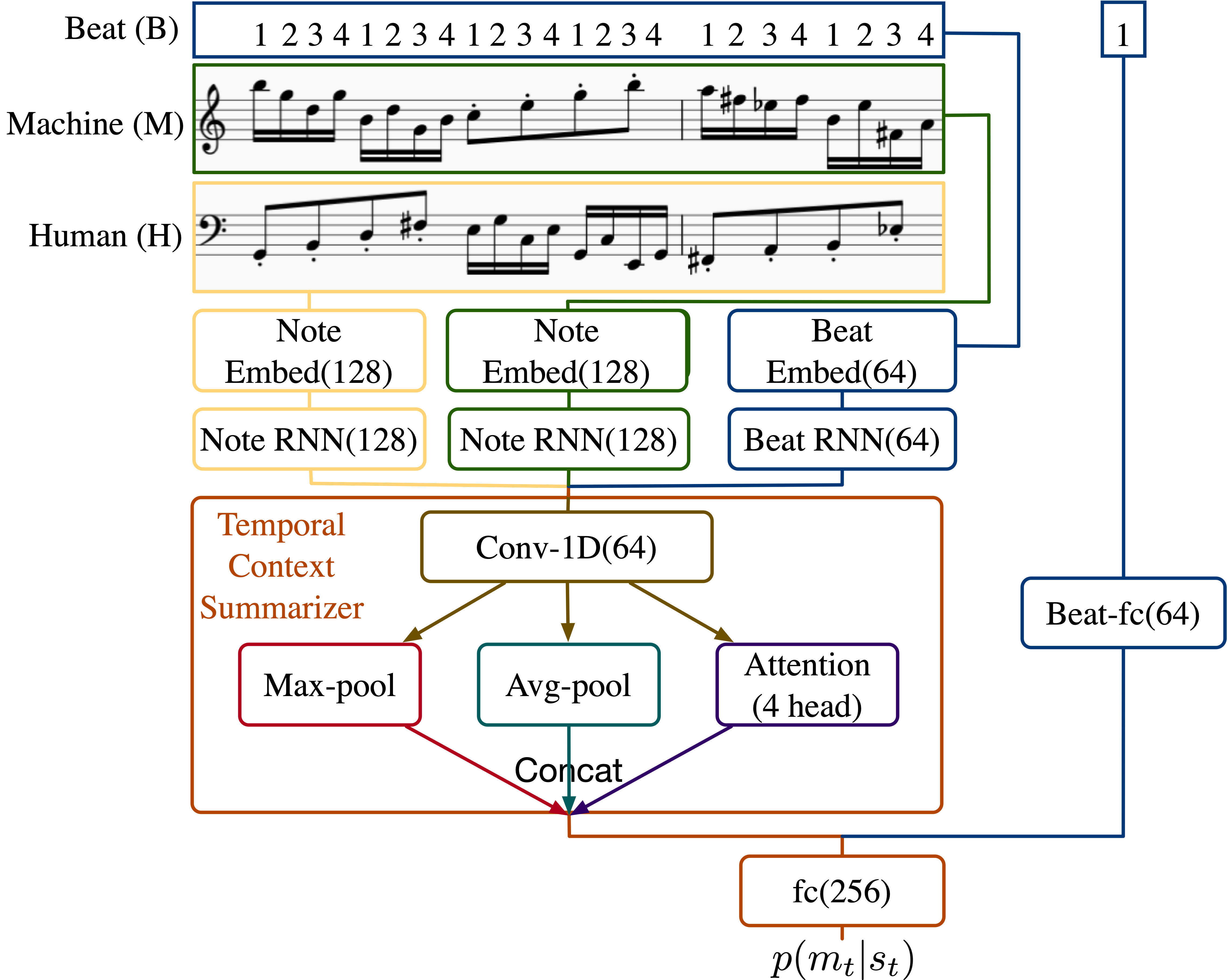}
	\caption{Model architecture of the generation model. The concatenation of the output of the human branch, machine branch and beat branch are fed into the Temporal Context Summarizer. Reward model (a) uses the same architecture, while reward models (b)(c)(d) use a similar architecture (see main text for details).}
	\label{fig:arch}
\end{figure}

\subsubsection{Model Specification}

Inspired by~\cite{bahdanau2015neural,rush2015neural,yin2016abcnn}, we design an attention-guided recurrent convolution neural network, which enhances CNN with a temporal attention mechanism. This attention mechanism enables the model to capture some aspects of musical structure such as repeated motifs. The model architecture for the generation and the reward model (a) is shown in Fig. \ref{fig:arch}. It uses embedding layers to encode the note tokens and the beat tokens, followed by bi-directional GRU layers to capture long-term dependencies. Afterwards, a Temporal Context Summarizer uses the pooling mechanism and the attention mechanism to adaptively aggregate the temporal information. At last, both the summarized context information and the feature of the current beat are concatenated to output a probability distribution. The architecture of the reward model (b)(c)(d) are almost the same as Fig. \ref{fig:arch}, except that their inputs to the Temporal Context Summarizer are different (shown in Fig \ref{fig:framework}), and their final \textit{fc} layer outputs the probability of a range of notes, from $m_t$ to $m_{t+\Delta}, \Delta=16$.

As introduced above, the reward agent of RL-Duet contains four types of reward models. We use six pre-trained reward models, three of which are of the type (a), trained for 20 epochs with learning rate of 0.005, 0.01 and 0.05, respectively. The remaining three reward models are of type (b)(c)(d), respectively, trained with learning rate 0.05 for 20 epochs. We choose to add more reward models of type (a), since it shares the same architecture with the generation model (see Fig. \ref{fig:framework}(a)). The generation model and reward model (a) use the ``multi-hold'' pitch representation as it alleviates the extreme imbalance of the amount of the hold symbol and pitch numbers in the ``single-hold'' representation. However, we still use the ``single-hold'' representation for reward models (b)(c)(d), since our empirical results show that the ``multi-hold'' representation will generate uncommon rhythms, such as a dotted eighth note followed by a sixteenth note in the generated music. We suggest that it may be because that in these cases, the model predicts a range of notes, from $m_t$ to $m_{t+\Delta}$ at the same time, making the evolution of the temporal patterns difficult to model.

\subsubsection{Training Process}
There are two training stages in RL-Duet. First, the reward models are trained with maximum likelihood estimation. And then, the generation model is trained with the reinforcement learning objective for 100K training duets, while all the reward models are only used for inference. The discount factor $\gamma = 0.5$. The $\lambda$ for the generalized advantage estimator is 1. The pretrained joint-modeling reward model (a) with learning rate 0.01 is used to initialize the weights of the generation model.

\subsection{Objective Evaluation}

\begin{figure*}[t]
    \centering
    \subcaptionbox{Pitch Count per Bar\label{fig:pc}}{\includegraphics[width=0.64\columnwidth]{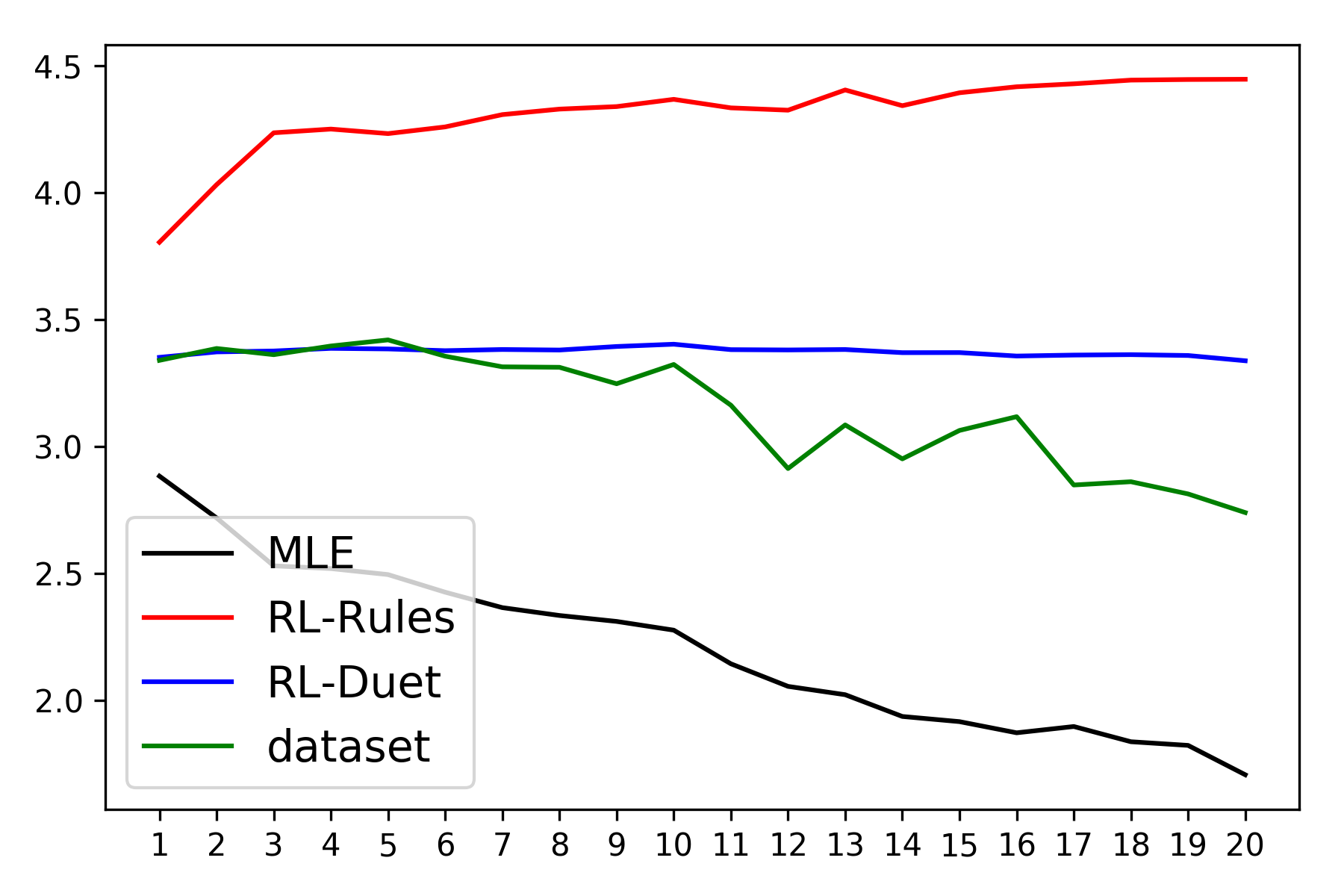}}
    \subcaptionbox{Pitch Interval\label{fig:pi}}{\includegraphics[width=0.64\columnwidth]{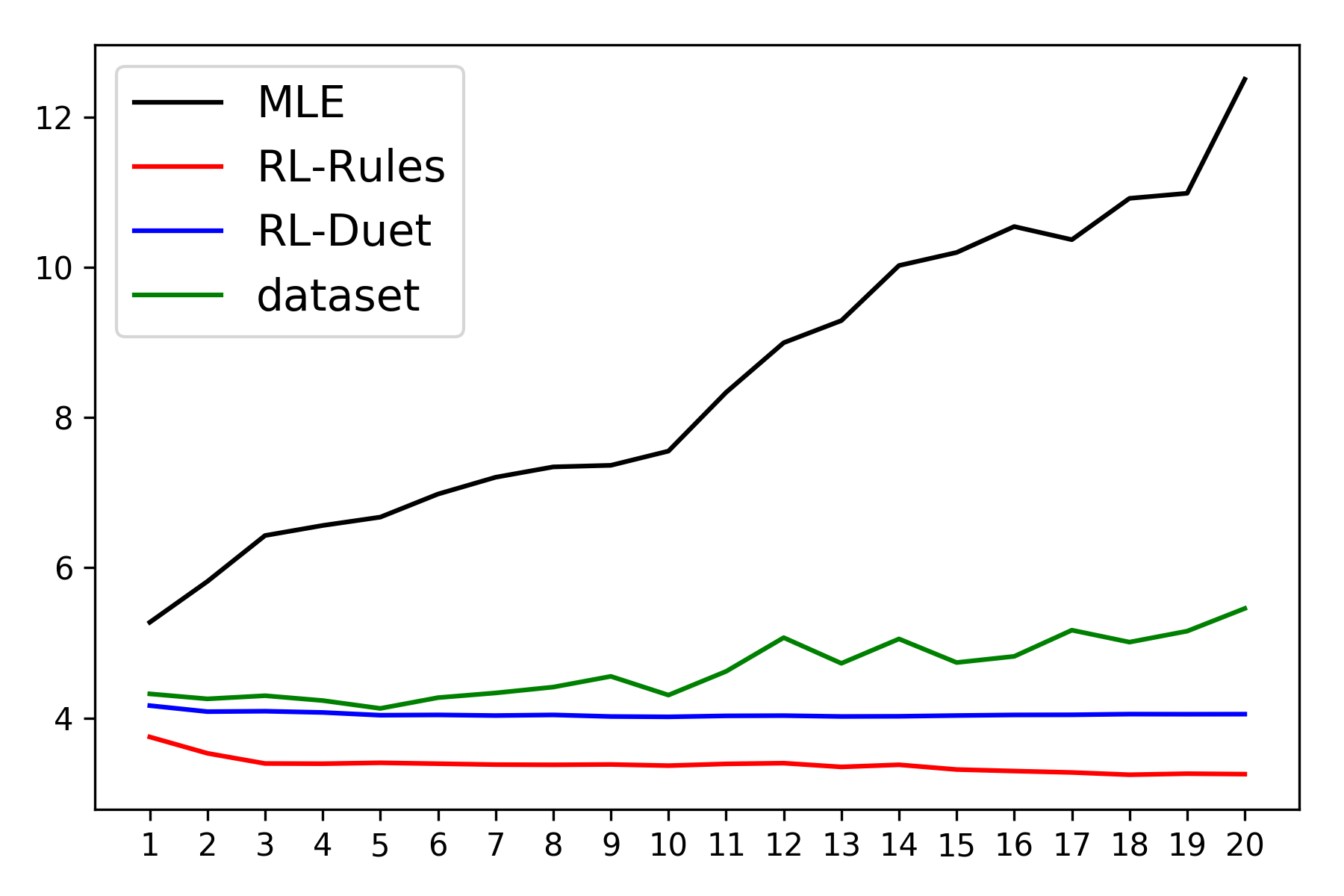}}
    \subcaptionbox{Inter-Onset-Interval\label{fig:ioi}}{\includegraphics[width=0.64\columnwidth]{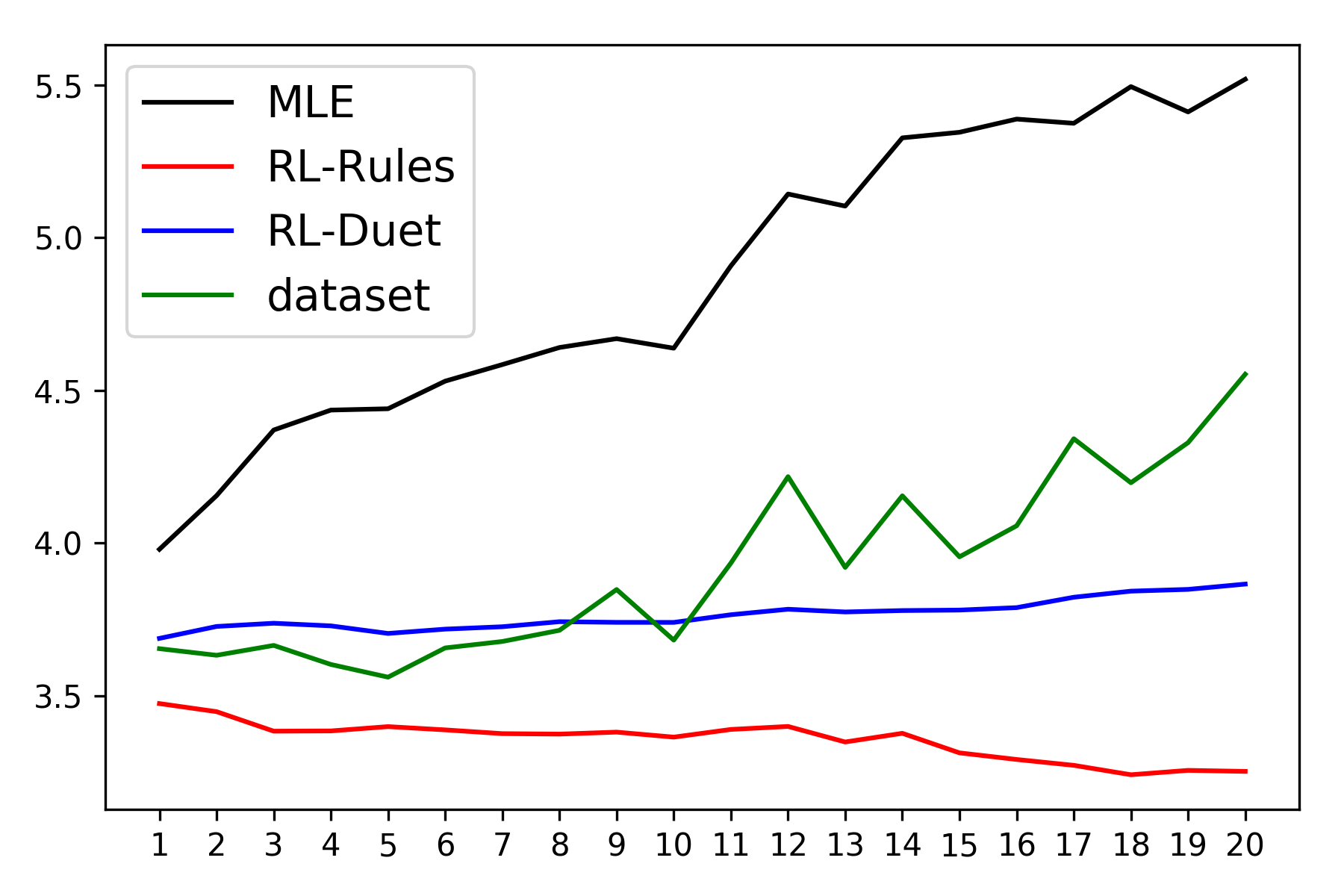}}
    \caption{Average objective metrics (vertical) versus measure indices (horizontal) across test pieces, showing how the generation quality changes over time within a music piece.}
	\label{fig:evolution}
\end{figure*}

In this section, we compared the performance of RL-Duet with a baseline MLE model and a rule-based reinforcement leaning model. The rule-based reinforcement leaning model is our implementation of SequenceTutor \cite{jaques2017sequence}. SequenceTutor uses a model-based reward along with many rule-based rewards, and the weights between them are fine-tuned. In our implementation, we use the pre-trained reward model (a) to provide the model-based reward, with some of the rule-based rewards from SequenceTutor which is applicable in our setting and the same weights. We do not use all of the rule-based rewards from it, since we have different pitch and beat representation from them and we do not transpose all the music to the same C-major key. Note that, the rules they designed only care about the harmony of the single machine generated part, and contain no information on the coherence between the human part and machine part.

We follow \cite{yang2018evaluation} to perform some objective evaluation. Specifically, we use pitch count per bar (PC/bar), average pitch interval (PI) and average inter-onset-interval (IOI). Table \ref{tab:objective} shows that these metrics of RL-Duet are closer to those of the test set than the baselines. We also calculate the histogram-like features, namely pitch class histogram (PCH) and note length histogram (NLH). We use the earth moving distance to measure the difference between the distributions of the histogram-like features from the generative models and those of the test set. As shown in Table \ref{tab:objective}, our proposed RL-Duet generates pieces that are more similar to the test set than both baselines in terms of PCH, and is competitive for NLH.

\begin{table}[]
\resizebox{\columnwidth}{!}{%
\begin{tabular}{l|c|c|c|c|c}
\hline
         & PC/bar       & PI           & IOI          & PCH    & NLH   \\ \hline
Dataset  & 3.25         & 4.57         & 3.84         & -      & -     \\ \hline
MLE      & -0.90  & +3.01 & +0.82 & 0.0067 & \textbf{0.039} \\
RL-Rules & +0.96  & -1.13 & -0.46 & 0.0079 & 0.043 \\
RL-Duet  & \textbf{+0.12}  & \textbf{-0.48} & \textbf{-0.09} & \textbf{0.0057} & 0.042 \\ \hline
\end{tabular}%
}
\caption{Objective evaluation results. We report the average PC/bar, PI, and IOI of pieces in the test dataset. For the three generation models, we report their differences from the test dataset. We also report the earth moving distance on PCH and NLH between the generated music and the test dataset. The smaller difference or distance suggests better style imitation of the dataset. \label{tab:objective}}
\end{table}

Considering that the first two measures of the machine part were always initialized with the ground-truth score, we observed that there exists some degeneration of the generated part by all the three algorithms in comparison. Therefore, we calculate how the objective metrics evolve over time through the generation of each piece using a sliding window in Figure \ref{fig:evolution}. The horizontal axis shows the measure indices. We take a one-sided 4-measure sliding window, \textit{i.e.}, the objective metrics corresponding to measure index 1 is calculated on Measures 1-4. Most of the duets in the test dataset are shorter than 15 measures, while only 20\% of duets in the test dataset are longer than 24 measures. Therefore, the evolution of the metrics is calculated from measure 1 to measure 20. RL-Duet is the most robust and achieves the closest evaluation metrics to the ground-truth. Interestingly, we also find that MLE captures the trend of these metrics of the dataset. However, due to the exposure bias (described in Motivation Section), generation problems tend to accumulate over time, leading to significant drifts and divergence of the objective metrics.
It is clear from Figure \ref{fig:pi} that, as the generation goes on, the MLE model is prone to repeating the same pitch, which accords with the MLE RNN's ``bad behaviors'' in SequenceTutor \cite{jaques2017sequence}.

\subsection{Subjective User Study}
\label{subject-eval}
Since music appreciation is very subjective and different people may have different opinions on the quality of music, we conduct a wide-range subjective user study. A simple survey on their music background is conducted beforehand:

\begin{enumerate}
	\item \emph{Have you learned to play a music instrument (no less than 5 hours a week) for more than 5 years in total in the past?
		-Yes -No}
	\item \emph{How much time do you spend in listening to classical music each week?\\
		- Less than 1 hour
		- Between 1 and 3 hours
		- Between 3 and 5 hours
		- More than 5 hours
	}\par
\end{enumerate}

\begin{figure}[t]
	\centering
	\includegraphics[width=0.96\columnwidth]{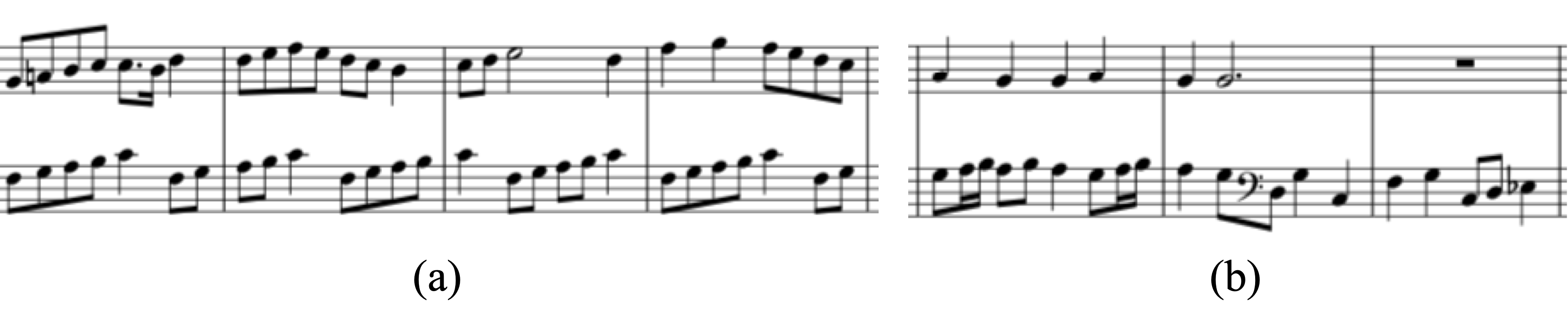}
	\caption{Examples showing typical problems of the rule-based model. The upper part is the human part, while the lower part is generated by RL-Rules. Problems: (a) Immediate repetition and misalignment between the repetitions and the rhythmic structure. (b) Large pitch leap.} 
	\label{fig:flaw}
\end{figure}

To perform the subjective evaluation, we compared the performance of RL-Duet with the baseline MLE model. The RL-Rules model is excluded since we observed some apparently bad generation of RL-Rules. Fig. \ref{fig:flaw} demonstrates immediate repetition, but with bizarre rhythmic structure, and large pitch leaps in the generated music of RL-Rules. The reason why rule-based model fails may be the inconsistency of the rule-based rewards. These rewards and the weights balancing them are carefully designed to adapt to the settings in SequenceTutor. However, when directly transplanting them in our setting, it easily fails.

We performed paired comparisons: Each subject was presented with a pair of two short duet excerpts (8 seconds) with the same human part. These excerpts were randomly truncated from generated duets, which were randomly chosen for each subject. After listening to each pair of duet samples, the participant was asked to choose which was preferred. Each pair of duet samples was evaluated by 20 subjects. 
125 subjects took the test and 2000 valid votes were collected. 19 of them (with 233 votes) had learned a musical instrument before, and 28 of them (with 617 votes) spent more than 1 hour listening to classical music per week. 

\begin{figure*}[t]
    \centering
    \subcaptionbox{Musical Instrument Skills\label{fig:subjective-instrument}}{\includegraphics[width=0.65\columnwidth]{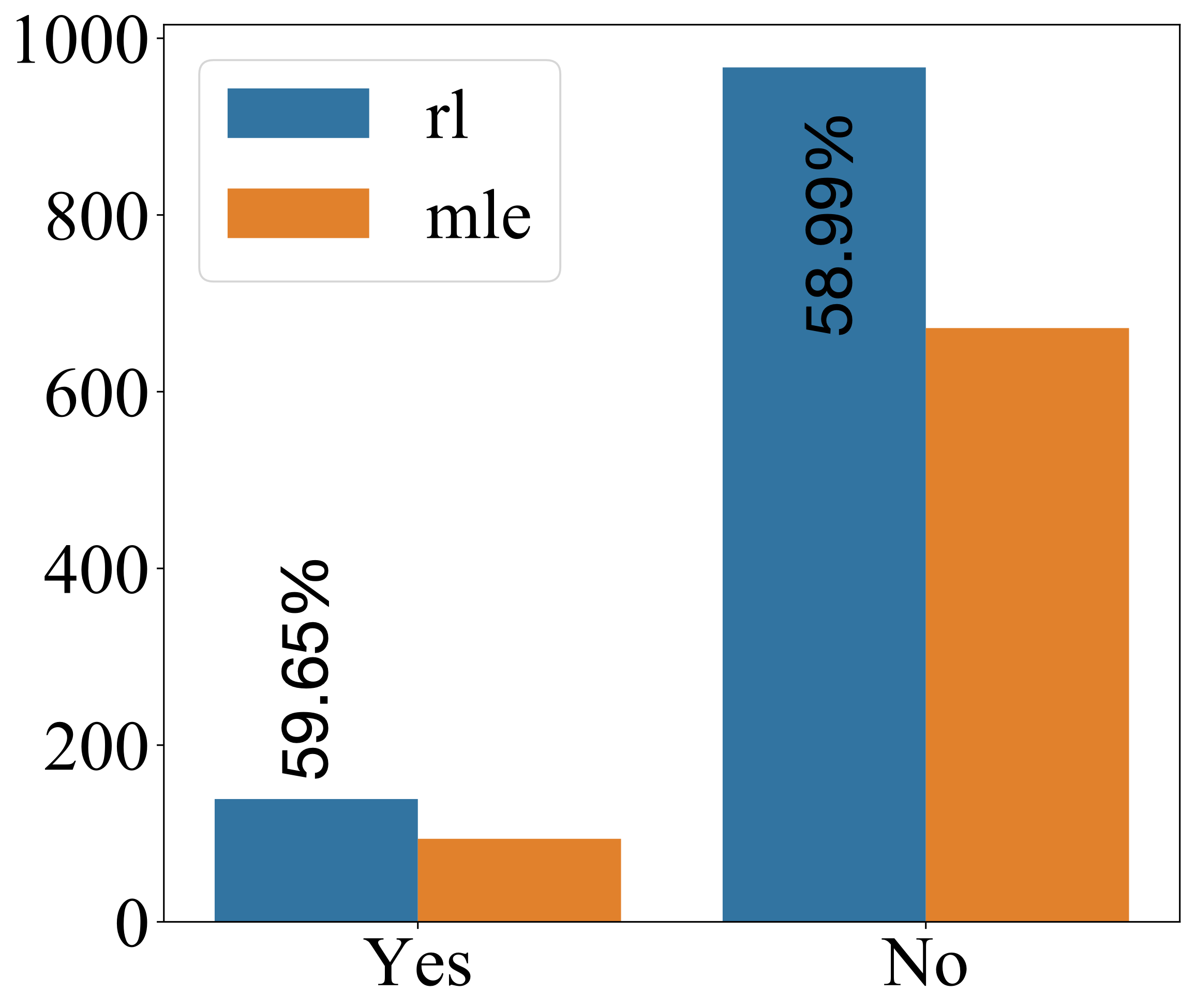}}
    \subcaptionbox{Classical Music Listening\label{fig:subjective-classic}}{\includegraphics[width=0.63\columnwidth]{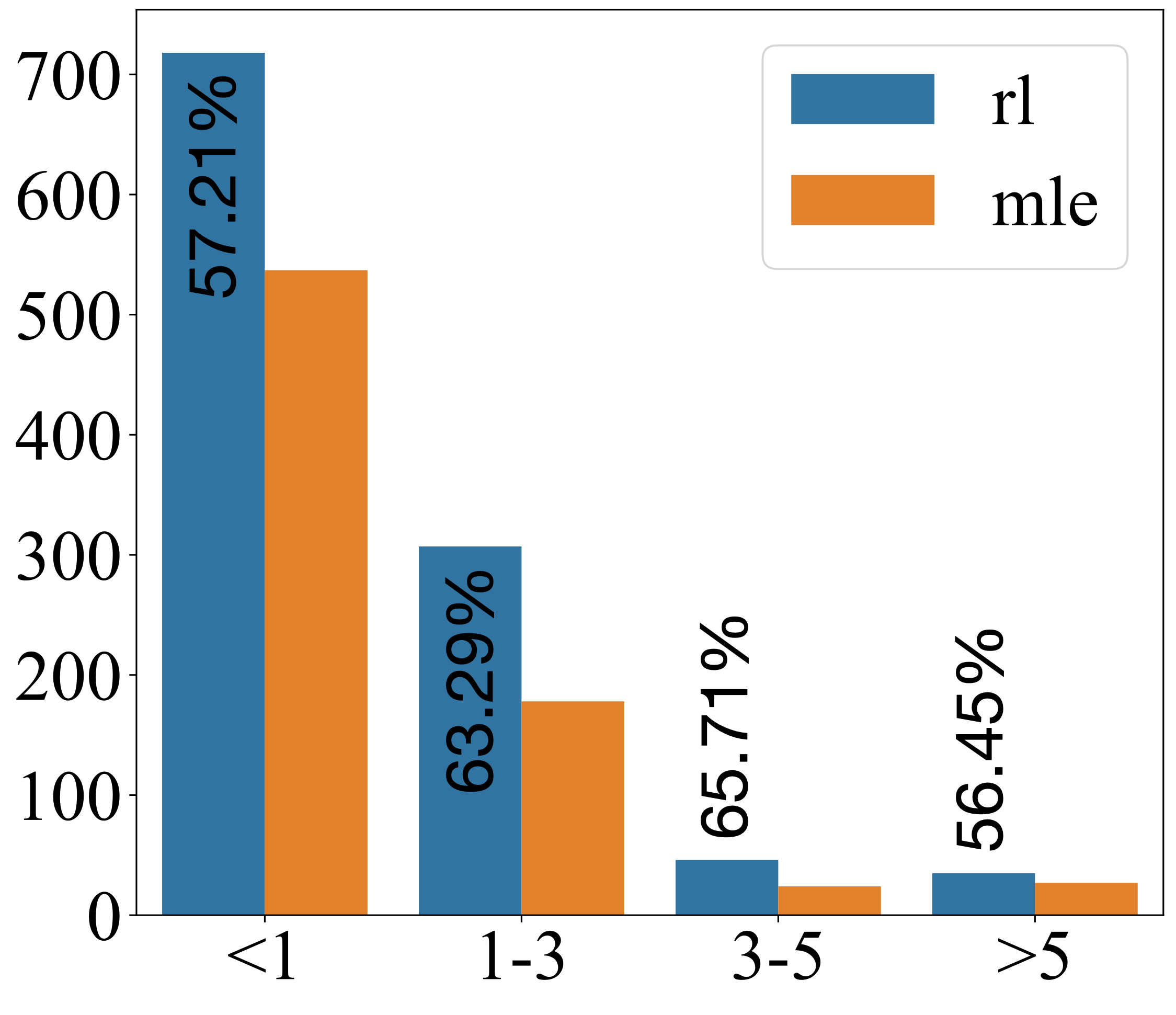}}
    \subcaptionbox{Summary\label{fig:subjective-instruments}}{\includegraphics[width=0.64\columnwidth]{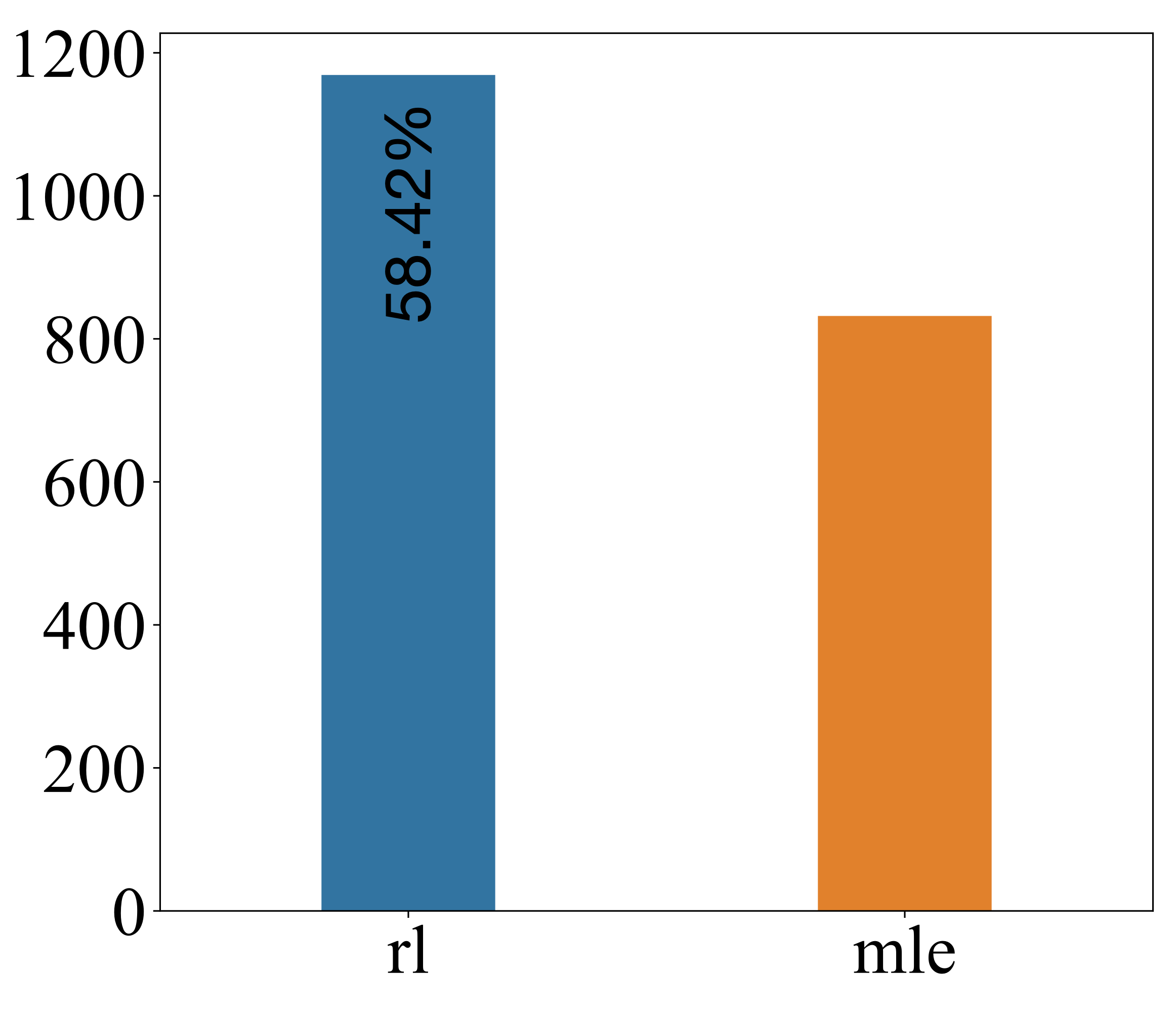}}
    \caption{Subjective evaluation results. The vertical axis is the number of votes. Blue bars and yellow bars mean the votes that prefer RL-Duet and the MLE model respectively. The subjects are classified into several groups according to their musical background: (a) with or without musical instrument skills, and (b) time spent in listening to the classical music per week. (c) The summary of the subjective evaluation. In total, 58.42\% votes prefer RL-Duet.}
	\label{fig:subjective}
\end{figure*}

Fig~\ref{fig:subjective} shows the paired comparison results. When the level of classical music familiarity is higher, the distinction between the RL-Duet and MLE model is more obvious. (An exception is the result of those who spent more than 5 hours per week listening to classical music, where too few votes are available and the results may be unreliable.) In total, around 58.42\% subjects prefer duets generated by RL-Duet to those generated by the MLE model. Considering that the paired two duets share the same human part, it can be hard to distinguish the two duets if the subject has little music background. For subjects with more performance and listening experiences, the preference of RL-Duet over MLE is more pronounced. 

\subsection{Interactive Generation}
\begin{figure}[ht!]
	\centering
	\includegraphics[width=1\columnwidth]{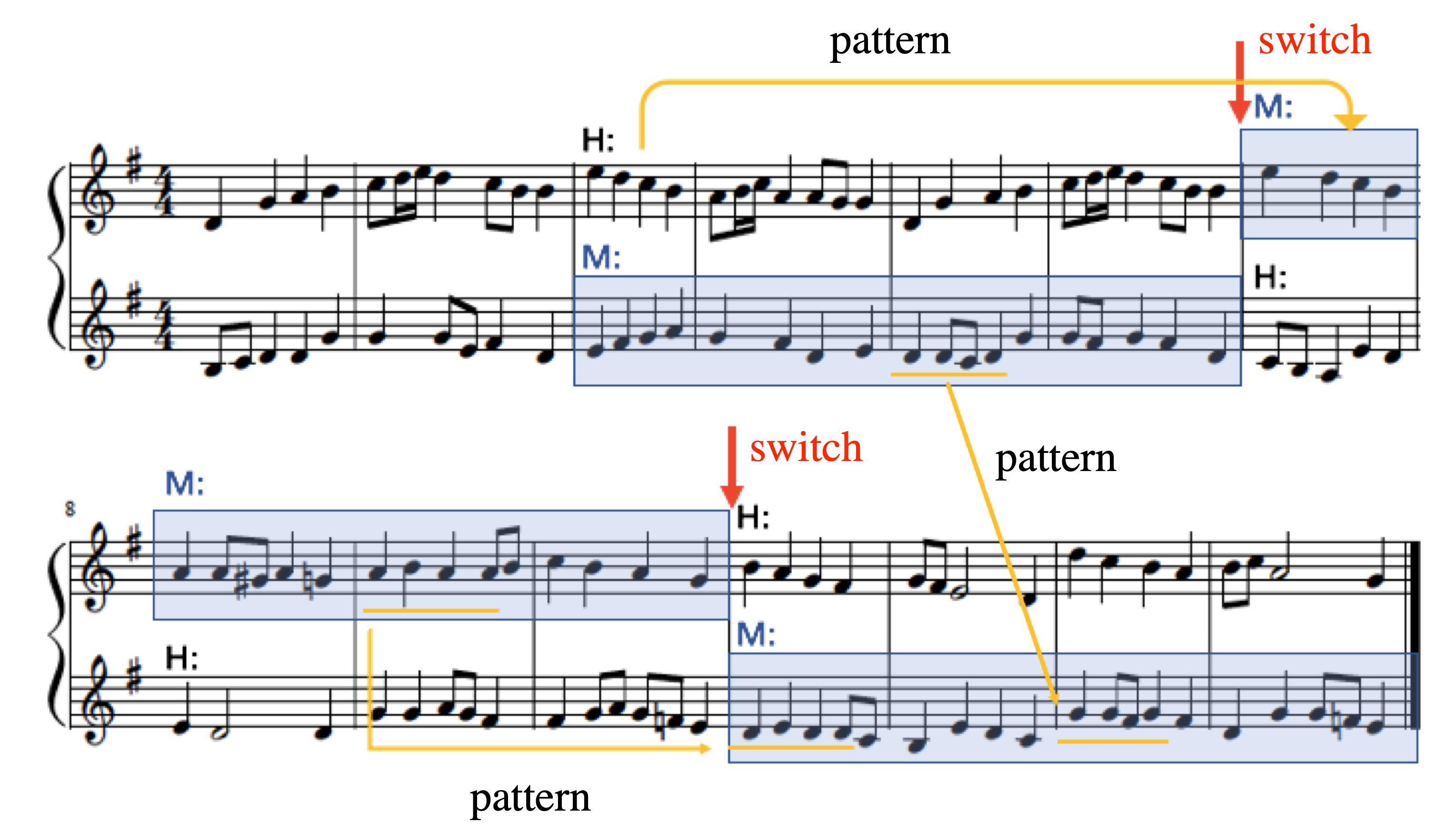}
	\caption{Interactive Generation. The first two measures of the machine part are given as the seed. The machine part generated by RL-Duet is covered with blue masks. Three motifs are highlighted with yellow arrows.}
	\label{fig:inter-generation}
\end{figure}

In previous sections, the roles of human and machine part are assumed to be fixed: the human part is always produced by the human, and the machine part is generated by RL-Duet. In the following, we extend RL-Duet to a more interactive setting, where the role of the human and the machine can be switched. Fig. \ref{fig:inter-generation} shows an example. Starting in Measure 6, the roles of the human and the machine are switched, and then in Measure 10, the roles are switched back. The human (machine) plays Part 1 (2) before the switch but plays Part 2 (1) after the switch. Even though the human part in this example is also a fixed part from the test data, which is invariant during generation, the machine part shows interesting responses to previous generated notes, by both the human and the machine. Fig. \ref{fig:inter-generation} shows three motifs in the machine part, in which the first relates to the human-generated part, while the other two relate to the machine part.

\section{Conclusion}

This paper presents RL-Duet, the first reinforcement learning model for online music accompaniment generation using an ensemble of reward models, in order to support real-time interactive music generation in a human-machine duet setup. Apart from the MLE model, RL-Duet generates music with better global coherence of the whole generated sequence. A comprehensive reward model considers the compatibility of the machine-generated notes with both the intra-part and inter-part contexts from the horizontal and the vertical views. Objective evaluation shows that RL-Duet has better style imitation of the dataset than an MLE baseline and a rule-based baseline. Subjective evaluation shows a higher preference on pieces generated by RL-Duet than those generated by the MLE baseline. For future work, we plan to integrate this model in an interactive human-machine duet improvisation system and investigate how the human musician interacts with the machine partner.

\section{ Acknowledgments}
This work is (jointly or partly) funded by the NSFC (Grant No. 61876095 and 61751308), Beijing Academy of Artificial Intelligence (BAAI), and National Science Foundation grant No. 1846184. We also thank Yujia Yan for the valuable discussions on music composition.

\bibliographystyle{aaai}
\bibliography{AAAI-JiangN.1016}

\end{document}